\documentclass[preprint,12pt]{elsarticle}

\usepackage{amssymb}

\usepackage{lineno}

\journal{}

\usepackage{hyperref}
\usepackage[export]{adjustbox}

\usepackage{xcolor}

\makeatletter
\def\ps@pprintTitle{%
	\let\@oddhead\@empty
	\let\@evenhead\@empty
	\def\@oddfoot{\textit{{\small Preprint version \ifx\@journal\@empty \else submitted to \@journal\fi\hfill \today}}}%
	\let\@evenfoot\@oddfoot}
\makeatother

\usepackage[procnames]{listings}
\usepackage{xcolor}

\begin{document}
	
	\definecolor{red}{RGB}{0,0,0}
	\definecolor{comments}{RGB}{125,125,125}
	\definecolor{black}{RGB}{0,0,0}
	
	\definecolor{string}{HTML}{226E6D}
	
	\definecolor{mygray}{rgb}{0.5,0.5,0.5}
	
	\lstset{language=Python, 
		basicstyle=\fontsize{8.7}{10}\ttfamily,
		keywordstyle=\color{black},
		commentstyle=\color{comments},
		stringstyle=\color{black},
		showstringspaces=false,
		identifierstyle=\color{black},
		procnamekeys={def,class},
		morekeywords={as,with},
		deletekeywords={compile}, 
		  numbers=left,                    
		  numbersep=5pt,                   
		  	numberstyle=\tiny\color{mygray},
		  	frame=single,
	framexleftmargin=20pt,
	framexrightmargin=5pt,
	framexbottommargin=5pt,
	framextopmargin=5pt,
		xleftmargin=5pt,
		frame=tblr,
		framesep=0pt,
		framerule=1pt,
		rulecolor=\color{gray},
		aboveskip=0pt,
		belowskip=5pt		 				
	}

\begin{frontmatter}




\title{InferPy: Probabilistic Modeling with Deep Neural Networks Made Easy\vspace{-16pt}}


\author[ual]{\vspace{-1pt}Javier C\'{o}zar}
\ead{jcozar87@ual.es}

\author[idsia]{Rafael  Caba\~{n}as}
\ead{rcabanas@idsia.ch}

\author[ual,cdtime]{Antonio Salmer\'{o}n}
\ead{antonio.salmeron@ual.es}

\author[ual,cdtime]{\\Andr\'{e}s R. Masegosa\vspace{-10pt}}
\ead{andresmasegosa@ual.es}

\address[ual]{Department of Mathematics, University of Almer\'{i}a, Spain \vspace{-4pt}}
\address[cdtime]{CDTIME, University of Almer\'{i}a, Spain \vspace{-4pt}}
\address[idsia]{Istituto Dalle Molle di Studi sull'Intelligenza Artificiale (IDSIA), Switzerland \vspace{-25pt}}

\begin{abstract}

\vspace{-20pt}
InferPy is a Python package for probabilistic modeling with deep neural networks.  It defines a user-friendly API that trades-off model complexity with ease of use, unlike other libraries whose focus is on dealing with very general probabilistic models at the cost of having a more complex API. In particular, this package allows to define, learn and evaluate general hierarchical probabilistic models containing deep neural networks in a compact and simple way. InferPy is built on top of Tensorflow Probability and Keras. 

\end{abstract}

\begin{keyword}
Deep Probabilistic modeling \sep Hierarchical probabilistic models\sep Neural networks \sep Bayesian layers \sep Tensorflow \sep User-friendly


\end{keyword}

\end{frontmatter}



\vspace{-18pt}
\section{Introduction}
\label{sec:intro}

\vspace{-12pt}
Advances in variational methods \cite{zhang2018advances}
have made possible the development of a new formalism, namely \textit{deep probabilistic modeling} \cite{masegosa2019probabilistic}, which combines probabilistic models within deep neural networks (DNNs) to capture complex non-linear relationships among random variables. The release of multiple libraries for deep probabilistic modeling \cite{tran2018simple,bingham2019pyro} are greatly expanding the adoption of these powerful probabilistic modeling techniques. 
However, these libraries 
are usually difficult to use, especially when defining  distributions  containing NNs, which requires dealing with multidimensional matrices (i.e. tensors).

This paper presents a new version of InferPy  as a high-level Python API for probabilistic modeling with  DNNs with a strong focus on ease of use. (A major released of the code has been performed: from 0.2.x to 1.2.x.) 
The main differences with the previous version \cite{cabanas2019inferpy}  are the following ones.  
Models can now contain DNNs to model non-linear relationships among random variables. The API has been significantly changed to make it compatible with the use of DNNs. Inferpy relies now on Tensorflow Probability (TFP) \cite{tran2018simple} (Inferpy's previous version relied on Edward \cite{tran2016edward}, which is deprecated).  

%

\vspace{-13pt}
\section{Background}
\label{}

\vspace{-12pt}
Probabilistic models with DNNs are usually found in the literature under the name of \textit{deep generative models} or \textit{Bayesian deep learning}
\citep{salakhutdinov2015learning}. The former focuses on density estimation using DNNs, while the latter focuses on treating the parameters of
a DNN as random variables to capture the model uncertainty. See \cite{masegosa2019probabilistic} for a recent review of  all these models. 
Along these lines, new software tools have appeared to deal with probabilistic models containing DNNs
\cite{cabanas2019inferpy,tran2018simple,bingham2019pyro}. These  tools usually fall under
the umbrella term \textit{probabilistic programming languages} (PPLs)
\citep{ghahramani2015probabilistic}, and provide support for methods for reasoning about complex
probabilistic models. Some examples are TFP \citep{tran2018simple}, Pyro \citep{bingham2019pyro}, etc.

%
%
%
%
%
%
%
%
%
%
%

\vspace{-13pt}
\section{Software Framework }
\vspace{-13pt}

%
%
%

The main features of InferPy are:
(i) Its simple API allows easy  prototyping of  probabilistic models including DNNs; 
(ii) Unlike TFP, it is not required to have a strong background in the inference methods available (Variational Inference
\cite{zhang2018advances, masegosa2019probabilistic}
and Monte Carlo methods \cite{brooks2011handbook}) as many details are hidden to the user;
(iii) InferPy runs seamlessly on CPUs and GPUs.
InferPy can be seen as an upper layer for working with TFP. Thus,  models that can be defined in InferPy are those that can be defined using TFP.
InferPy is distributed as open-software (Apache-2.0)  using Pypi and its source code is available at GitHub (see Tab. \ref{tab:exe} and \ref{tab:src}).



\vspace{-13pt}
\section{Illustrative Example}
\vspace{-12pt}

\begin{figure}[h!bt]
	\centering
\includegraphics[scale=0.97]{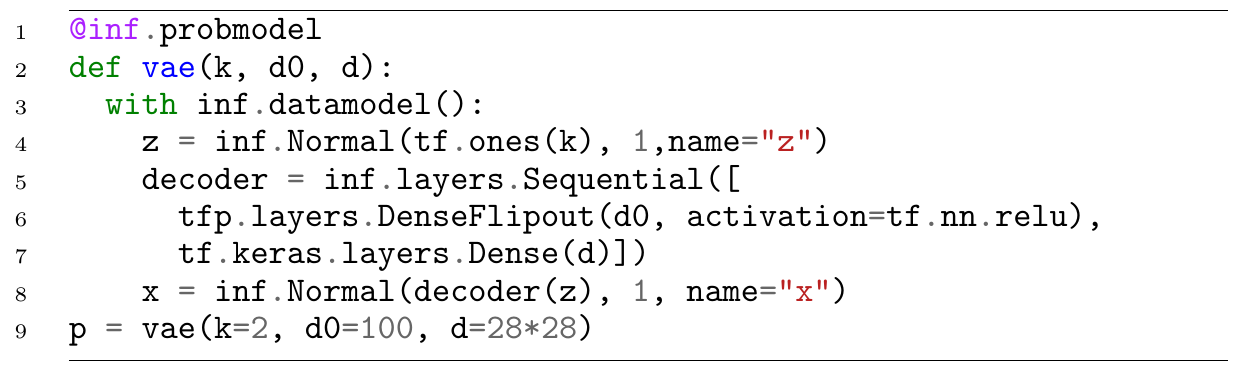}
\vspace{-10pt}
	\caption{p-model and decoder}
	\label{fig:code_pmodel}
	\vspace{0pt}
\end{figure}

For illustrating the usage of InferPy,  we will consider a variational autoencoder (VAE) \cite{doersch2016tutorial}, as it is one of the most widely used probabilistic models containing deep NNs. 
In a VAE, every object has a unknown latent representation (a code), modeled with a multivariate Gaussian (a distribution over possible codes). This latent representation gives rise to a multivariate Gaussian distribution over the observed representation (the decoded observation of this object) by passing the latent representation through a NN called the \textit{decoder}. This part of the model is defined in Fig.~\ref{fig:code_pmodel} (lines 1 to 8) and the creation of an instance (line 9), which is an object of class \texttt{inf.models.probmodel}. A probabilistic model in InferPy is defined as a function with the decorator \texttt{@inf.probmodel}. Following the statistical inference terminology, we refer to this part of the model as the $p$ model.

Random variables are objects of class \texttt{inf.models.RandomVariable}. Variables
composing a probabilistic model are those instantiated during the execution of its decorated function. The  \texttt{with inf.datamodel()} syntaxis is used to indicate the InferPy variables contained within this construct are replicated for every data sample. Every replicated variable is conditionally independent given the previous random variables (if any) defined outside this \texttt{with} statement\footnote{In contrast to other libraries, the number of replications will be automatically calculated just before the inference.}. This construct enormously simplify the code of the model.

%
%
%

\begin{figure}[h!bt]
	\centering
	\includegraphics[scale=0.97]{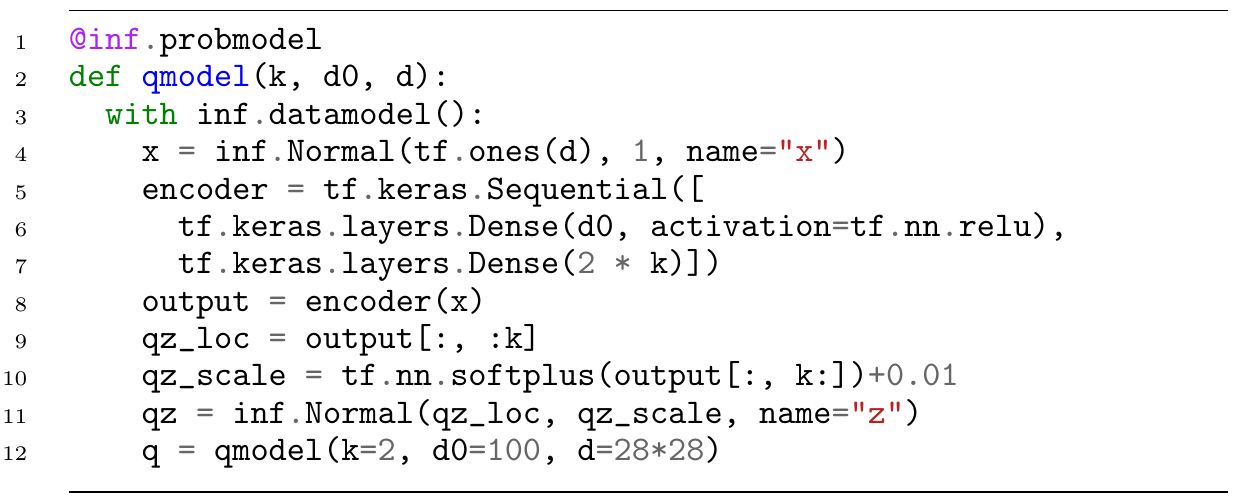}
	\vspace{-10pt}
	\caption{q-model and encoder}
	\label{fig:code_qmodel}
	\vspace{0pt}
\end{figure}

The \textit{encoder} part of a VAE defines the inference part of the model: given the observed  representation of an object we need to find the posterior probability over possible latent representations (codes) of this object, in the form a multivarate Gaussian. In a VAE, this inference part is defined using an \textit{amortized variational inference} \cite{zhang2018advances, masegosa2019probabilistic} scheme, which relies on an \textit{encoder network}. Following the variational inference terminology, we call this part of the model as the $q$ model. As shown in Fig.~\ref{fig:code_qmodel}, this part is similarly defined with the same decorator. The correspondence between the variables in the \textit{decoder} part and the \textit{encoder} part of the model is done by the argument \texttt{name}, i.e., they should be the same. For the NNs definition, standard  Keras code can be used. However, in case of Bayesian NNs (e.g., \texttt{tfp.layers.DenseFlipout} defines a \textit{Bayesian layer}),  the \texttt{Sequential} model provided by InferPy must be used. This provides a Bayesian treatment of the \textit{decoder network}.

%
%
%
%

\begin{figure}[h!bt]
	\centering
	\includegraphics[scale=0.97]{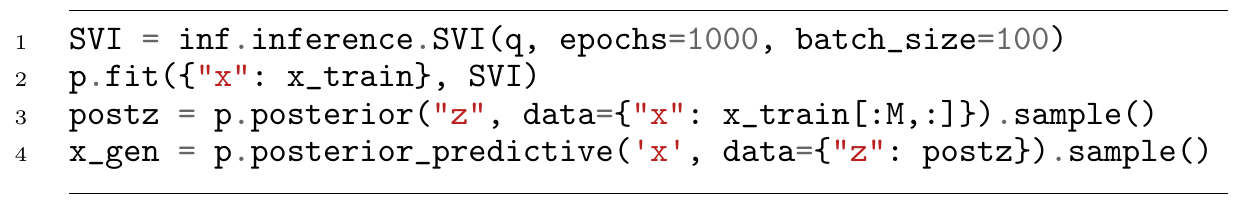}
	\vspace{-10pt}
\caption{Inference from the posterior distributions}
\label{fig:code_post}
	\vspace{0pt}
\end{figure}

 A minimal example using (stochastic) variational inference \cite{zhang2018advances, masegosa2019probabilistic} as a learning engine is given in Fig.~\ref{fig:code_post}. Even though the learning algorithm can be further configured, in this case, an object of class \texttt{inf.inference.SVI}  is created with the q-model, the \texttt{epochs} (number of iterations) and \texttt{batch\_size} as input arguments. The optimization starts when the method \texttt{fit()} is invoked.
Finally, we might sample from the posterior of \texttt{z} (latent representation) or from the posterior predictive (generating new samples). 

%
%

The analogous TFP code for this model is far more complex. This can be found in the online documentation (see Tab. \ref{tab:exe} and \ref{tab:src}), together with other examples and a complete user manual.


\vspace{-15pt}
\section{Conclusions}
\label{}
\vspace{-10pt}
InferPy can now represent probabilistic models containing DNNs using a simple and compact API. As most of the inference details are hidden, this package can be used for users without a strong probabilistic background.

\vspace{-15pt}
\section*{Acknowledgements}

\label{}
\vspace{-16pt}
  Authors have been jointly supported by the Spanish Ministry of Science, Innovation and Universities and by the FEDER under the projects TIN2015-74368-JIN, and TIN2016-77902-C3-3-P.



\vspace{-20pt}
\bibliographystyle{elsarticle-num} 
\bibliography{neuro-inferpy-deep}


\clearpage
\section*{Metadata}
\label{}

\section*{Current executable software version}
\label{}

\begin{table}[!h]
	\resizebox{\linewidth}{!}{
		\begin{tabular}{|l|p{5.5cm}|p{8.8cm}|}
			\hline
			\textbf{Nr.} & \textbf{(executable) Software metadata description} &\\
			\hline
			S1 & Current software version & 1.3.0 \\
			\hline
			S2 & Permanent link to executables of this version  &  https://pypi.org/project/inferpy/1.3.0/\\
			\hline
			S3 & Legal Software License & Apache 2.0 \\
			\hline
			S4 & Computing platform/Operating System & Linux, OS X, Microsoft Windows, Unix-like \\
			\hline
			S5 & Installation requirements \& dependencies & Pip, Python 3.5-3.7, tensorflow 1.12.1-1.15.0
			tensorflow-probability 0.7.0, networkx 2.2.0-3.0\\
			\hline
			S6 &  Link to user manual  &  https://inferpy.readthedocs.io/\\
			\hline
			S7 & Support email for questions & inferpy.api@gmail.com \\
			\hline
		\end{tabular}
	}
	\caption{Software metadata}
	\label{tab:exe} 
\end{table}

\section*{Current code version}
\label{}

\begin{table}[!h]
	\resizebox{\linewidth}{!}{
		\begin{tabular}{|l|p{5.5cm}|p{8.8cm}|}
			\hline
			\textbf{Nr.} & \textbf{Code metadata description} & \\
			\hline
			C1 & Current code version & 1.3.0 \\
			\hline
			C2 & Permanent link to code/repository used of this code version & \footnotesize https://github.com/PGM-Lab/InferPy/tree/1.3.0\\
			\hline
			C3 & Legal Code License   & Apache 2.0 \\
			\hline
			C4 & Code versioning system used & github \\
			\hline
			C5 & Software code languages, tools, and services used & Python\\
			\hline
			C6 & Compilation requirements, operating environments &  Python 3.5-3.7, tensorflow 1.12.1-1.15.0
			tensorflow-probability 0.7.0, networkx 2.2.0-3.0\\
			\hline
			C7 &Link to developer documentation/manual &  https://inferpy.readthedocs.io/\\
			\hline
			C8 & Support email for questions & inferpy.api@gmail.com\\
			\hline
		\end{tabular}
	}
	\caption{Code metadata}
	\label{tab:src} 
\end{table}

\end{document}